\documentclass[10pt,letterpaper]{article}\usepackage[]{graphicx}\usepackage[]{color}
\pdfoutput=1

\makeatletter
\def\maxwidth{ %
  \ifdim\Gin@nat@width>\linewidth
    \linewidth
  \else
    \Gin@nat@width
  \fi
}
\makeatother

\definecolor{fgcolor}{rgb}{0.345, 0.345, 0.345}

\usepackage{framed}
\makeatletter
 {\par\unskip\endMakeFramed%
 \at@end@of@kframe}
\makeatother

\definecolor{shadecolor}{rgb}{.97, .97, .97}
\definecolor{messagecolor}{rgb}{0, 0, 0}
\definecolor{warningcolor}{rgb}{1, 0, 1}
\definecolor{errorcolor}{rgb}{1, 0, 0}
\newenvironment{knitrout}{}{} 

\usepackage{alltt}

\usepackage{cogsci}
\usepackage{pslatex}
\usepackage{apacite}
\usepackage{gb4e}
\usepackage{amsmath}
\usepackage{amsfonts}

\title{Feature overwriting as a finite mixture process: \\ 
Evidence from comprehension data}
 
 \author{{\large \bf Shravan Vasishth (vasishth@uni-potsdam.de)} \\
 Department of Linguistics, University of Potsdam, Germany, and \\
Centre de Recherche en Math\'ematiques de la D\'ecision, CNRS, UMR 7534, Universit\'e Paris-Dauphine,\\ PSL Research University, Paris, France.\\
  \AND {\large \bf Lena J\"ager (lena.jaeger@uni-potsdam.de)} \\
 Department of Linguistics, University of Potsdam, Germany.\\  
  \AND {\large \bf Bruno Nicenboim (bruno.nicenboim@uni-potsdam.de)} \\
 Department of Linguistics, University of Potsdam, Germany.\\  
 }
\IfFileExists{upquote.sty}{\usepackage{upquote}}{}
\begin{document}

\maketitle

\begin{abstract}
The ungrammatical sentence \textit{The key to the cabinets are on the table} is known to lead to an illusion of grammaticality.  As discussed in the meta-analysis by J\"ager et al., 2017, faster reading times are observed at the verb \textit{are} in the agreement-attraction sentence above compared to the equally ungrammatical sentence \textit{The key to the cabinet are on the table}. 
One explanation for this facilitation effect is the feature percolation account: the plural feature on \textit{cabinets} percolates up to the head noun \textit{key}, leading to the illusion.  
An alternative account is in terms of cue-based retrieval account (Lewis \& Vasishth, 2005), which assumes that the non-subject noun \textit{cabinets} is misretrieved due to a partial feature-match when a dependency completion process at the auxiliary initiates a memory access for a subject with plural marking.
We present evidence for yet another explanation for the observed facilitation. Because the second sentence has two nouns with identical number, it is possible that these are, in some proportion of trials, more difficult to keep distinct, leading to slower reading times at the verb in the first sentence above; this is the feature overwriting account of Nairne, 1990. We show that the feature overwriting proposal can be implemented as a finite mixture process.
We reanalysed ten published data-sets, fitting hierarchical Bayesian mixture models to these data assuming a two-mixture distribution. We show that in nine out of the ten studies, a mixture distribution corresponding to feature overwriting furnishes a superior fit over both the feature percolation and the cue-based retrieval accounts. 
\end{abstract}

\section{Introduction}

It is well-known that sentences such as (\ref{example1}) 
can lead to an illusion of grammaticality.  The sentence is
ungrammatical because of the lack of number agreement between
the subject \textit{key} and the auxiliary \textit{are}.
Note that the second noun, \textit{cabinets}, and the auxiliary \textit{are} agree in number, but no syntactic agreement is possible between these two elements.

\begin{exe} 
\ex
\begin{xlist}
\item \label{example1}
The key to the cabinets are on the table.
\item \label{example2}
The key to the cabinet are on the table.
\end{xlist}
\end{exe}


Many sentence comprehension studies have shown that
 the illusion has the effect that
 the auxiliary \textit{are} is read faster in (\ref{example1}) compared to the equally ungrammatical sentence (\ref{example2}) (see \citeNP{JaegerEngelmannVasishth2017} for a review). In contrast to (\ref{example1}), in (\ref{example2}) the second noun (\textit{cabinet}) is singular and does not agree with the auxilary in number.


Several explanations have been proposed for the illusion of grammaticality in (\ref{example1}) vs.\ (\ref{example2}).
We discuss two of these here.
The feature percolation account proposes that in (\ref{example1}) the plural feature on \textit{cabinets} can, in some proportion of trials, move or percolate up to the head noun \textit{key} (see \citeNP{patson2016misinterpretations} for recent evidence for this model). The head noun now has the plural feature, leading to an illusion of grammaticality compared to (\ref{example2}), where no such feature percolation occurs. 
Another prominent explanation, due to \citeA{wagersetal}, is the retrieval interference account. Here, in ungrammatical sentences like (\ref{example1}), a singular verb would be predicted; but when the plural verb \textit{are} is encountered, a cue-based retrieval process \cite{lewisvasishth:cogsci05} is triggered: The verb triggers an access (called a retrieval) for a noun that is plural marked and is a subject. A parallel cue-based associative memory access leads to the retrieval of a partially matching noun in memory (\textit{cabinets}) that agrees in number but is not the subject. This partial match leads to a successful retrieval and an illusion of grammaticality.\footnote{The cue-based retrieval account may a priori be implausible because it predicts that an incorrect dependency is built between \textit{cabinets} and \textit{are}; building such a dependency would imply that the sentence has the implausible meaning that the cabinets are on the table. The reader should detect such an implausible meaning and this should lead to a slowdown rather than facilitation.} 

As we show next, there is evidence for both these accounts: a facilitatory effect is generally present in the published data. 

\subsection{The facilitatory effect in reading time in the ``illusion of grammaticality'' data-sets}

We first establish that a facilitatory effect is found in studies comparing sentences like (\ref{example1}) and (\ref{example2}).
In connection with the meta-analysis relating to studies on cue-based retrieval reported in \citeA{JaegerEngelmannVasishth2017}, we had obtained the raw data from 10 studies on sentences like (\ref{example1}) and (\ref{example2}). These were reading-time studies reported in \citeA{Dillon-EtAl-2013}, \citeA{lago2015agreement}, and \citeA{wagersetal}. Except for the eyetracking experiment by Dillon and colleagues, all the other studies were self-paced reading experiments. In these data-sets, the dependent measure was reading time in milliseconds at the auxiliary or the region following it (most of the 10 studies found statistically significant effects in this post-critical region).

We first reanalyzed these 10 data-sets in order to confirm the facilitatory effect reported.\footnote{The published studies had other experimental conditions that we do not discuss here. The published studies also used a trimming procedure to analyze the data, and their analysis was done on the raw millisecond scale. Thus, our analysis has some differences from the original analyses, but the conclusions are substantially unchanged.} We fit Bayesian hierarchical models to each data-set using Stan \cite{stan-manual:2016}. 
We fit Bayesian models
because of the ease with which statistical models can be defined flexibly to reflect the cognitive process of interest. 

The model specification was as follows. 
Assume that (i) $i$ indexes participants, $i=1,\dots,I$ and $j$ indexes items, $j=1,\dots,J$; 
(ii) $y_{ij}$ is the reading time in milliseconds for the $i$-th participant reading the $j$-th item;
 and 
 (iii) the predictor $x$, which represents the experimental manipulation, is sum-coded ($\pm 1$). In our case, the condition (\ref{example1}) is coded $+1$ and the condition (\ref{example2}) is coded $-1$.
 
 Then,  the data $y_{ij}$ (reading times in milliseconds) are defined to be generated by the following process:

\begin{equation} \label{eq:lmm1}
y_{ij} \sim LogNormal(\beta_1+ \beta_2 x_{ij} + u_i + w_j, \sigma_e^2)
\end{equation}

\noindent
where $u_i \sim Normal(0,\sigma_u^2)$, $w_j \sim Normal(0,\sigma_w^2)$ and $\sigma_e^2$ is the error variance. The terms $u_i$ and $w_j$ are called varying intercepts for participants and items respectively; they represent by-subject and by-item adjustments to the fixed-effect intercept $\beta_1$. The variances $\sigma_u^2$ and $\sigma_w^2$ represent between-participant (respectively item) variance.\footnote{This so-called crossed participants and items varying intercepts linear mixed model can be made more complex by adding varying slopes for the factor $X$ by participant and by item, but for space reasons we do not consider these more complex models here.} 
The facilitation effect is the estimate of $\beta_2$ (on the log scale).

As priors, we chose the Cauchy(0,2.5) distribution for all coefficients, and a half-Cauchy (with only positive values) for the standard deviations. This are mildly informative priors \cite{Gelman14}
which express the belief that that the most likely value of the parameter is near 0, but allows for a wide range of non-zero values because of the fat tails of the Cauchy.

\begin{figure}[!htbp]
\centering
\begin{knitrout}
\definecolor{shadecolor}{rgb}{0.969, 0.969, 0.969}\color{fgcolor}
\includegraphics[width=\maxwidth]{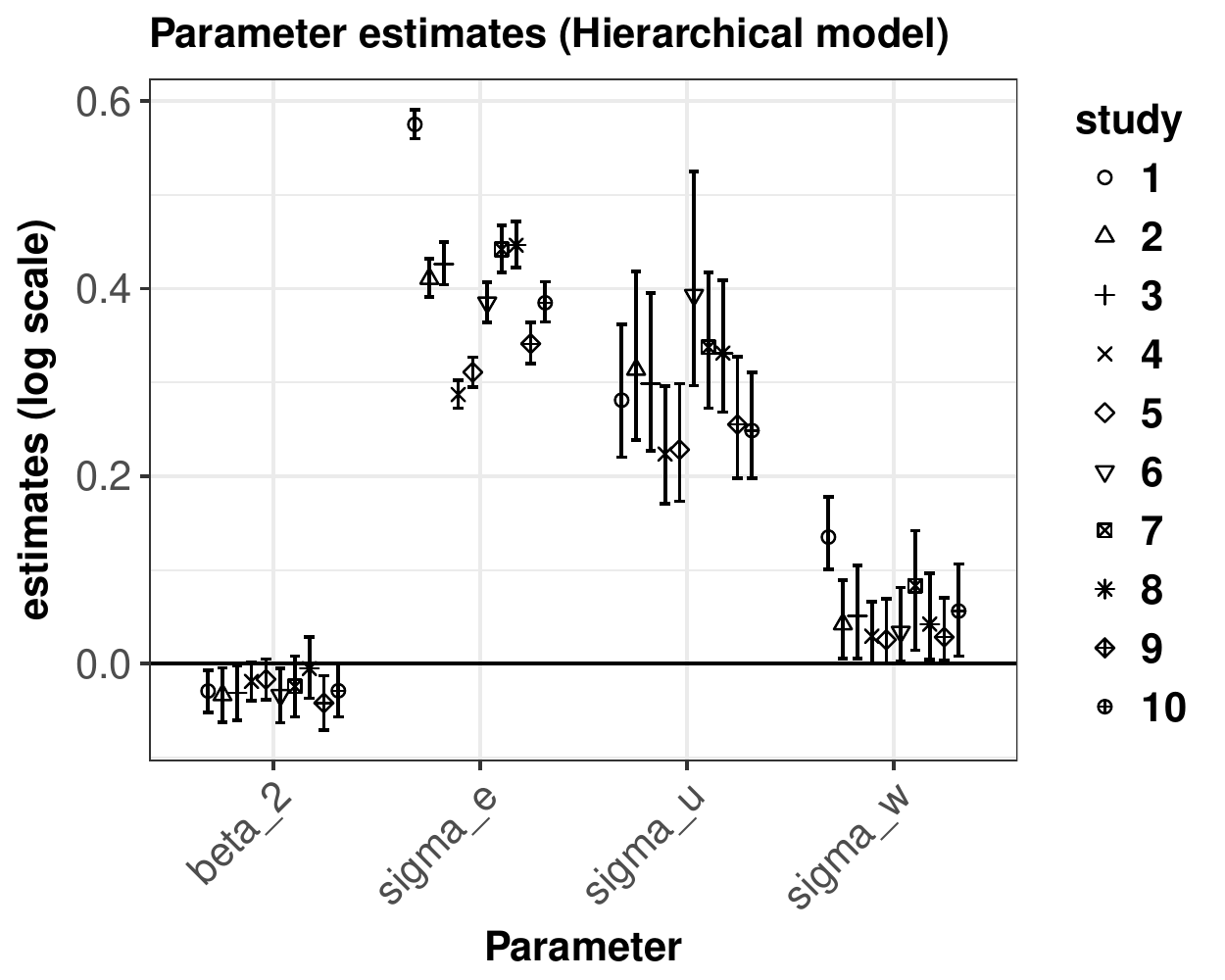} 

\end{knitrout}
\caption{The parameter estimates of the hierarchical model fitted to the 10 data-sets. 
The condition representing (\ref{example1}) is coded $+1$ and the condition representing (\ref{example2}) is coded $-1$, so that parameter \texttt{beta\_2} shows a facilitation effect if its value is negative.
Shown are the estimates of the facilitatory effect (\texttt{beta\_2}), and the standard deviations of (i) the error (\texttt{sigma\_e}), (ii) the by-subjects varying intercepts (\texttt{sigma\_u}), and (iii) the by-items varying intercepts (\texttt{sigma\_w}).} \label{fig:LMMs}
\end{figure}

As shown in Figure~\ref{fig:LMMs}, the effects in each study consistently show negative estimates of $\beta_2$, which indicates a facilitation in reading time at the auxiliary or a subsequent region.
This is consistent with both the feature percolation and retrieval interference accounts. 
There is a third explanation for the observed facilitation effect  in these studies, which we turn to next.

\subsubsection{An alternative explanation for the facilitatory effect}

Consider the ungrammatical example sentences again. These are repeated below for convenience:

\begin{exe} 
\ex
\begin{xlist} 
\item \label{example1rep} 
The key to the cabinets are on the table.
\item \label{example2rep}
The key to the cabinet are on the table.
\end{xlist}
\end{exe}

In example~(\ref{example2rep}),
both the nouns are marked singular, whereas in example~(\ref{example1rep}) the nouns have different number marking. 
As discussed in \citeA{VillataFranck},
the similarity in number of the two nouns in (\ref{example2rep}) could be the underlying cause for increased processing difficulty, compared to (\ref{example1rep}).
The identical number marking in (\ref{example2rep}) could lead to increased confusability between the two nouns, leading to longer reading times at the moment when a subject noun is to be accessed at the auxiliary verb. 
The feature overwriting model of \citeA{nairne1990feature} formalizes this idea. To quote (p.\ 252):
\textit{An individual feature of a primary memory trace is assumed to be overwritten, with probability $F$, if that feature is matched in a subsequently occurring event. Interference occurs on a feature-by-feature basis, so that, if feature $b$ matches feature $a$, the latter will be lost with probability $F$}.

The Nairne proposal has a natural interpretation as a 
 finite mixture process. Specifically, feature overwriting could occur with a higher probability in example~(\ref{example2rep}) compared to (\ref{example1rep}). This assumption implies that the reading times in both (\ref{example2rep}) and (\ref{example1rep}) are generated from a mixture of two distributions. 
In a particular trial, if no feature overwriting occurs, the reading time would come from a Lognormal distribution with some location and scale parameters; this situation would result in 
minimal processing difficulty in  carrying out a retrieval and detecting the ungrammaticality. In other trials,
when feature overwriting does occur, 
the reading time would have a larger location parameter, and possibly also a larger scale parameter; this would represent the cases where additional difficulty occurred due to feature overwriting. 

An explicit assumption here is that feature overwriting could occur in both (\ref{example2rep}) and (\ref{example1rep}), but the proportion would be higher in (\ref{example2rep}). It is also possible to assume that feature overwriting only occurs in 
(\ref{example2rep}), but due to space reasons we do not consider this and other alternative models here. 

Thus, in the mixture model implementation of the Nairne proposal, one distribution will have a larger location parameter (and perhaps also the scale parameter).
In the modelling presented below, one goal is to estimate the mixing proportions of these distributions.
In the results section, we will refer to the proportion of the slow reading time distributions in (\ref{example2rep}) as \texttt{prob\_hi}, and in (\ref{example1rep}) \texttt{prob\_lo}. The suffixes \texttt{hi} and \texttt{lo} here refer to whether we expect confusability to be high or low.

To summarize, the feature percolation, cue-based retrieval, and 
feature overwriting models all predict facilitation in the ungrammatical sentences (\ref{example1rep}) compared to (\ref{example2rep}), but the underlying generative process assumed in each model is different. Feature percolation and feature overwriting can be seen as finite mixture models of different types, and cue-based retrieval can be seen as implemented by the standard hierarchical model.
Our goal here is to implement all the three proposals as statistical models and then compare their relative fit to the data in order to adjudicate between them.
Before we do this, we introduce finite mixture models.

\section{Finite mixture models}

A finite mixture model assumes that the independently distributed outcome 
$y_i, i=1,\dots,N$
is drawn from
one of several distributions. Each distribution's identity is controlled by a Categorical distribution.
For example, assume that we have $K$ distributions with location parameter (the mean) $\mu_k \in \mathbb{R}$ and scales (standard deviation) $\sigma_k \in (0,\infty)$, where $k=1,\dots,K$. 
Assume also that we have a vector of probabilities $<\lambda_1,\dots,\lambda_K>=\Lambda$ 
that represent the mixing proportions.
The parameters $\lambda_k$ are non-negative values and they sum to 1. 

Thus, if the $K$ distributions are  mixed in proportion $\Lambda$, where $\lambda_k \geq 0$ and $\sum_{k=1}^K \lambda_k = 1$, for each outcome $y_i$ there is a latent variable $z_i \in \{1,\dots,K\}$ with a Categorical distribution\footnote{The Categorical distribution can be seen here as the Bernoulli distribution in the case where K=2. In this paper, we focus only on the K=2 case.} parameterized by $\lambda: z_i \sim \hbox{\texttt{Categorical}}(\lambda)$.  The variable $y_i$ is then distributed as follows:

\begin{equation}
y_i \sim Normal(\mu_{z_i},\sigma_{z_i}^2)
\end{equation}

Assuming that each of the $K$ mixture distributions $f(\cdot)$ has a vector of parameters $\theta_k$ associated with it, the mixture density can be written in the following manner:

\begin{equation}
p(y_i\mid \theta,\Lambda) = \lambda_1 \cdot f(y_i\mid \theta_1)+\dots+\lambda_K \cdot f(y_i\mid \theta_K)
\end{equation}

A random variable $Y$ with the above density can then be written in abbreviated form as follows \cite{fruhwirth2006finite}.

\begin{equation}
Y \sim \lambda_1 f(y\mid \theta_1)+\dots+ \lambda_K f(y\mid \theta_K)
\end{equation}

In this paper, we consider a mixture of LogNormals with $K=2$; we choose LogNormals to model reading times because reading times must be greater than $0$ and follow a LogNormal distribution. We will write the models as follows:

\begin{equation}
\begin{split}
Y \sim& \lambda_1 \cdot LogNormal(\mu_1+\delta,\sigma_{1}^2)+(1-\lambda_1) \cdot LogNormal(\mu_1,\sigma_2^2)\\
~& \hbox{where } \sigma_1^2 = \sigma_2^2\hbox{ or } \sigma_1^2 \neq \sigma_2^2\\
\end{split}
\end{equation}

\noindent
The parameter $\delta$ marks the shift in the mean in the first mixture distribution relative to the second mixture distribution.
Note that the scale parameters ($\sigma_1, \sigma_2$) can be either identical (homogeneous variances) in both distributions, or different (heterogeneous variances). 
We will consider both types of models here.

The above models assume that the data are independent. When we have repeated measures data, the independence assumption is no longer valid. 
In order to address this issue, finite mixture models can be made hierarchical by adding varying intercepts for subjects (indexed by $i$) and items (indexed by $j$):

\begin{equation}
\begin{split}
y_{ij} \sim& \lambda_1 \cdot LogNormal(\mu_1+\delta+u_i+w_j,\sigma_{1}^2)+\\
           ~& (1-\lambda_1) \cdot LogNormal(\mu_1+u_i+w_j,\sigma_2^2)\\
\end{split}
\end{equation}

\noindent
where $u_i \sim Normal(0,\sigma_u^2)$ and $w_j \sim Normal(0,\sigma_w^2)$. Thus, the mixture model with $K=2$ will have the following parameters: 
four variance components, $\sigma_{1}^2, \sigma_2^2, \sigma_u^2$, 
and $\sigma_w^2$; two coefficients $\mu_1$ and $\delta$; and 
two probabilities $\lambda_1$ and $\lambda_2=(1-\lambda_1)$. 


\section{An evaluation of the Nairne feature overwriting proposal}

\subsection{Method}

\subsubsection{Implementing the Nairne proposal}

We fit the homogeneous and heterogeneous variance hierarchical mixture models to the 10 reading time data-sets that compared reading times at the auxiliary or the following region for sentences like (\ref{example1rep}) and (\ref{example2rep}). 

The data were assumed to be generated from a two-mixture Lognormal distribution with either a homogeneous variance in both mixture distributions, or heterogeneous variances. 
Thus, for the high confusability condition (\ref{example2rep}), we considered two models:

\begin{equation}
\begin{split}
       ~&\hbox{\underline{Homogeneous variance feature overwriting model}}\\
y_{ij} \sim& \hbox{prob\_hi} \cdot LogNormal(\beta+\delta+u_i+w_j,\sigma_{e}^2)+\\
           ~& (1-\hbox{prob\_hi}) \cdot LogNormal(\beta+u_i+w_j,\sigma_e^2)\\
           ~& \hbox{where: }\\
           ~& u_i \sim Normal(0,\sigma_u^2), w_k \sim Normal(0,\sigma_w^2)\\ 
\end{split}
\end{equation}

\begin{equation}
\begin{split}
       ~&\hbox{\underline{Heterogeneous variance feature overwriting model}}\\
y_{ij}
\sim& \hbox{prob\_hi} \cdot LogNormal(\beta+\delta+u_i+w_j,\sigma_{e'}^2)+\\
           ~& (1-\hbox{prob\_hi}) \cdot LogNormal(\beta+u_i+w_j,\sigma_e^2)\\
           ~& \hbox{where: }\\
           ~& u_i \sim Normal(0,\sigma_u^2), w_k \sim Normal(0,\sigma_w^2)\\ 
\end{split}
\end{equation}

\noindent
In both models, 
$y_{ij}$ 
is the reading time in milliseconds from subject $i$ and item $j$.  The probability  \texttt{prob\_hi} represents the mixing probability of the distribution that generates the slow reading times corresponding to trials where feature overwriting occurred (\ref{example2rep}). Although not shown, another mixture distribution is defined for example (\ref{example1rep}); here, 
\texttt{prob\_lo} represents the mixing probability of the distribution that generates the slower reading times corresponding to the trials where feature overwriting occurred. 

The homogeneous variance model assumes that both mixture distributions have the same standard deviation $\sigma_e$.  The heterogeneous mixture model assumes that the mixture distribution that leads to the slower reading times is assumed to have both a different mean ($\beta+\delta$) and a different standard deviation ($\sigma_{e'}$) than the other distribution.
Alternative models can be fit which relax these assumptions, but due to space constraints we consider only these two models.

We had the following priors for the parameters:

\begin{equation}
\begin{split}
\hbox{prob\_hi} \sim& Beta(1,1)\\
\beta, \delta ~& \sim Cauchy(0,2.5)\\
\sigma_e, \sigma_{e'}, \sigma_u, \sigma_w ~& \sim Cauchy(0,2.5)\\
         ~& \hbox{constraint: } \sigma_e, \sigma_{e'}, \sigma_{u}, \sigma_{w}>0\\
\end{split}
\end{equation}

The priors for the variance components (the standard deviations $\sigma_e$, $\sigma_{e'}$, $\sigma_u$, $\sigma_w$) and the coefficients representing the means of the Lognormal distributions ($\beta, \delta$) are  mildly informative priors, as in the standard hierarchical model above. These Cauchy priors assume that values of the parameters near 0 are the most likely ones, but extreme values are possible. 
The Beta(1,1) prior for the mixing probabilities 
expresses a large prior uncertainty, and express the assumption that the probability is equally likely to be any value between 0 and 1.

\subsubsection{Baseline models}

As baselines, we fit a model corresponding to the retrieval interference account (the standard hierarchical model shown in equation~\ref{eq:lmm1} and  summarized in Figure~\ref{fig:LMMs}),  and the 
feature percolation proposal. The latter also assumes a mixture distribution, but only for the condition corresponding to  example (\ref{example1rep}). Recall that the claim is that in ungrammatical sentences, in some proportion of trials the plural feature on the distractor \textit{cabinets} moves up to the head noun. In (\ref{example2rep}), no such mixture process should occur because percolation never occurs; hence a standard hierarchical LogNormal distribution can be assumed here. We therefore defined the following generative process for (\ref{example1rep}):

\begin{equation}
\begin{split}
       ~&\hbox{\underline{Feature percolation model}}\\
y_{ij} \sim& \hbox{prob\_perc} \cdot LogNormal(\beta+\gamma+u_i+w_j,\sigma_{e}^2)+\\
           ~& (1-\hbox{prob\_perc}) \cdot LogNormal(\beta+u_i+w_j,\sigma_e^2)\\
           ~& \hbox{where: }\\
           ~& u_i \sim Normal(0,\sigma_u^2), w_k \sim Normal(0,\sigma_w^2), \gamma < 0 \\ 
\end{split}
\end{equation}

\noindent
Note that in the specification above the parameter $\gamma$, which represents the change in the location parameter, is constrained in the model to be negative; this is because the assumption in the feature percolation proposal is that percolation leads to faster reading time.

For sentences like (\ref{example2rep}), in which no percolation is assumed to occur, we simply assumed a LogNormal generative process:

\begin{equation}
y_{ij} \sim LogNormal(\beta+u_i+w_j,\sigma_{e}^2)
\end{equation}

\subsubsection{Model comparison}
Having fitted the homogeneous and heterogeneous variance models, as well as the baseline models (the cue-based retrieval and feature percolation models),
we need a method for comparing the quality of fit of the mixture models relative to the standard hierarchical models.
We use an approximation of the leave-one-out cross-validation (LOO), as discussed in 
\citeA{vehtari2016LOOwaic}. We find this approach attractive because it focuses on the predictive performance of the model. 
LOO compares the expected predictive performance of alternative models by subsetting the data into a training set (for estimating parameters) by excluding one observation. The difference between the predicted and observed held-out value can then be used to quantify model quality by successively holding out each observation. The sum of the expected log pointwise predictive density, $\widehat{elpd}$, can be used as a measure of predictive accuracy, and the difference between the $\widehat{elpd}$'s of competing models can be computed, including the standard deviation of the sampling distribution of the difference in $\widehat{elpd}$.
When comparing a model M1 with another model M2, if M2 has a higher $\widehat{elpd}$, then it has a better predictive performance compared to M1. The model comparisons are transitive; if a third model M3 has a higher $\widehat{elpd}$ than M2, then it has a better performance than M1 as well. 
Vehtari and colleagues have developed an efficient computation of LOO using Pareto-smoothed importance sampling (PSIS-LOO), This is what we use here. For details of PSIS-LOO, see \citeA{vehtari2016LOOwaic}.

\subsection{Results}

\begin{table*}[!htbp]
\begin{center}
\begin{tabular}{ccccccc}
      & \multicolumn{2}{c}{(a) Standard HLM vs.} & \multicolumn{2}{c}{(b) Percolation vs.} & \multicolumn{2}{c}{(c) Homogeneous variance vs.}\\
      & \multicolumn{2}{c}{Homogeneous variance} & \multicolumn{2}{c}{Homogeneous variance} & \multicolumn{2}{c}{Heterogeneous variance}\\
            & \multicolumn{2}{c}{mixture model} & \multicolumn{2}{c}{mixture model} & \multicolumn{2}{c}{mixture model}\\
Study & elpd\_diff & SE &    elpd\_diff & SE & elpd\_diff & SE \\ 
   1 & -0.29 & 1.67     &    29.55 & 6.97    & 0.57 & 1.09 \\ 
  2 & 56.98 & 13.57     &    76.34 & 14.26   & 15.20 & 6.07 \\ 
   3 & 97.62 & 16.10    &    112.40 & 17.43  & 57.12 & 11.11 \\ 
  4 & 71.29 & 14.08     &    84.78 & 14.12   & 19.66 & 8.77 \\ 
   5 & 112.74 & 18.17   &    120.45 & 18.56  & 63.28 & 18.12 \\ 
  6 & 66.84 & 12.59     &    85.97 & 13.88   & 43.58 & 12.18 \\ 
   7 & 72.45 & 13.76    &    80.93 & 14.72   & 80.92 & 14.41 \\ 
  8 & 88.50 & 14.60     &    90.22 & 14.77   & 40.17 & 11.87 \\ 
  9 & 78.35 & 14.21     &    108.10 & 16.04  & 26.21 & 7.76 \\ 
  10 & 90.08 & 14.14    &   105.23 & 15.02   & 33.59 & 11.95 \\ 
\end{tabular}
\end{center}
\caption{Comparison of the 10 sets of hierarchical models using PSIS-LOO. Shown are the differences in $\widehat{elpd}$ between (a) the standard hierarchical model and the homogeneous variance mixture model; (b) the feature percolation model and the homogeneous variance mixture model; and (c) the homogeneous vs.\ heterogeneous variance mixture model. Also shown are standard errors for each comparison. If the difference in $\widehat{elpd}$ is positive, this is evidence in favour of the second model. The pairwise model comparisons are transitive. 
These comparisons show that the heterogeneous variance mixture model has the best predictive performance.}\label{tab:allcomparisons}
\end{table*}

\begin{figure*}[!htbp]
\begin{knitrout}
\definecolor{shadecolor}{rgb}{0.969, 0.969, 0.969}\color{fgcolor}
\includegraphics[width=\maxwidth]{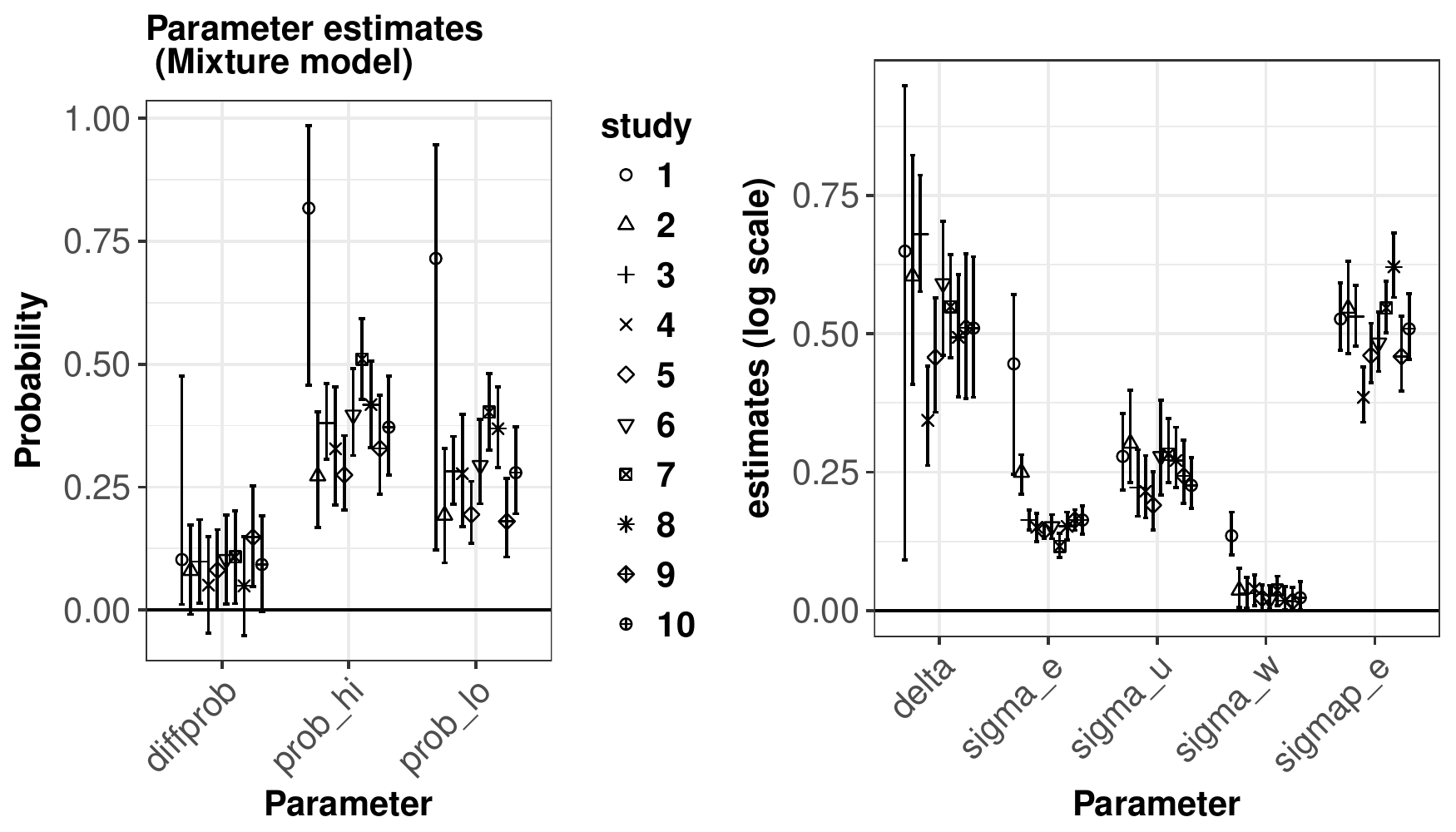} 

\end{knitrout}
\caption{Parameter estimates for the heterogeneous variance hierarchical mixture models.}\label{fig:HierMMs}
\end{figure*}

Table~\ref{tab:allcomparisons}  shows
model comparisons between 
the standard hierarchical model, corresponding to the retrieval interference account, and the homogeneous variance model.
The table shows that 
apart from study 1, the homogeneous variance feature overwriting model is clearly superior to the retrieval interference model because it has higher $\widehat{elpd}$ values. 
Table~\ref{tab:allcomparisons} also shows that the homogeneous variance feature overwriting model furnishes a better fit than the feature percolation model.
Finally, the table shows that, except for study 1, the heterogeneous variance model is superior to the homogeneous variance model.

Since the model comparisons are transitive, we can conclude that, among the models compared, the heterogeneous variance feature overwriting model characterises the data best. 
We therefore focus on the parameter estimates of the heterogeneous variance model below.
The estimates from the models for the 10 data-sets are shown in Figure~\ref{fig:HierMMs}. 
In this model, two noteworthy points are the following:
(i) The variance  of the high confusability distribution (\texttt{sigmap\_e}; this corresponds to $\sigma_{e'}$ in the models defined earlier) is relatively large compared to the other variance components;
(ii)
The difference in probabilities of the two mixture distributions, 
\texttt{diffprob}, is generally greater than 0 across all the studies; however, the uncertainty in the estimate of the probability in study 1 is very high. 
These two observations suggest that there is more variability in the reading time when the feature overwriting occurs, and that there some evidence that the proportion of trials with feature overwriting is higher in the condition with two singular nouns, consistent with the Nairne proposal.

In summary, overall there is good motivation to assume that in the
condition with two singular nouns (example~\ref{example2rep}), a proportion of trials comes from a distribution with a larger mean and larger standard deviation, and this proportion is higher than in the condition with one singular and one plural noun (example~\ref{example2rep}).

\subsection{Discussion}

We implemented as a statistical model the proposal that 
nouns with similar feature marking (here, number) may be more confusable due to feature overwriting in some proportion of trials, which in turn leads to occasional increase in difficulty in accessing the correct noun when a dependency is to be completed between the subject and the verb. By fitting Bayesian hierarchical two-mixture models, we showed that 9 out of the 10 data-sets showed evidence for this increased confusability in one condition over the other. 
The feature overwriting account for the ungrammatical sentences (\ref{example1rep}, \ref{example2rep}) appears to be superior to both the retrieval interference and feature percolation accounts. 

The three accounts make the same predictions for ungrammatical sentences---a facilitation effect. The modelling presented here allows us to quantitatively compare the relative fit of these proposals for these otherwise indistinguishable accounts. An interesting future direction is to evaluate the predictions of these models for grammatical sentences such as those considered in \citeA{franck2015task,VillataFranck}. We plan to address this in future work.


\section{Acknowledgments}

Our thanks to Brian Dillon, Sol Lago, Colin Phillips, and Matt Wagers for generously sharing their data.
We also benefitted a great deal from discussions with Julie Franck, Whitney Tabor, Aki Vehtari, and Michael Betancourt.
For partial support of this research,  we thank the Volkswagen Foundation through grant 89 953 to the first author.

\bibliographystyle{apacite}

\setlength{\bibleftmargin}{.125in}
\setlength{\bibindent}{-\bibleftmargin}

\bibliography{VasishthEtAlICCM2017.bib}


\end{document}